\RequirePackage{fix-cm}
\documentclass[twocolumn]{svjour3_arxiv}          
\smartqed  
\usepackage{graphicx}
\usepackage{subfigure}
\usepackage{color}
\usepackage{amsmath}
\usepackage{amssymb}
\usepackage{wrapfig}
\usepackage{algorithm2e}
\usepackage{pgfplots}
\usepackage{tikz}
\usepackage{url}
\usepackage{placeins}
\usepackage{txfonts}

\DeclareMathAlphabet{\mathcal}{OMS}{cmsy}{m}{n}
\newcommand{\R}{\mathbb{R}}
\newcommand{\Argmin}[1]{\arg~ \underset{#1}{\textrm{min}}}

\DeclareMathOperator{\rk}{rk}

\DeclareMathOperator{\Gr}{Gr}
\DeclareMathOperator{\St}{St}

\definecolor{Diagram1}{HTML}{101010}
\definecolor{Diagram2}{HTML}{003330}
\definecolor{Diagram3}{HTML}{3E6100}
\definecolor{Diagram4}{HTML}{0A8200}
\definecolor{Diagram5}{HTML}{DE7C01}
\definecolor{Diagram6}{HTML}{D90016}
\definecolor{Diagram7}{HTML}{8C00FF}

\journalname{}
\begin{document}

\title{pROST : A Smoothed $\ell_p$-norm Robust Online Subspace Tracking Method for Realtime Background Subtraction in Video
}


\author{Florian Seidel \and Clemens Hage \and Martin Kleinsteuber
}

\institute{Florian Seidel \at
Department of Informatics\\
Technische Universit\"{a}t M\"{u}nchen\\
Boltzmannstr. 3, 85748 Garching, Germany \\
\email{seidel.florian@gmail.com}\vspace{12pt}\\
Clemens Hage, Martin Kleinsteuber \at
Department of Electrical Engineering and Information Technology\\
Technische Universit\"{a}t M\"{u}nchen\\
Arcisstr.~21, 80333 Munich, Germany \\
\email{\{hage,kleinsteuber\}@tum.de}
}

\date{\today}

\maketitle
\begin{abstract}
An increasing number of methods for background subtraction use Robust PCA to identify sparse foreground objects. While many algorithms use the $\ell_1$-norm as a convex relaxation of the ideal sparsifying function, we approach the problem with a smoothed $\ell_p$-norm and present pROST, a method for robust online subspace tracking. The algorithm is based on alternating minimization on manifolds. Implemented on a graphics processing unit it achieves realtime performance. Experimental results on a state-of-the-art benchmark for background subtraction on real-world video data indicate that the method succeeds at a broad variety of background subtraction scenarios, and it outperforms competing approaches when video quality is deteriorated by camera jitter.
\end{abstract}
\section{Introduction}
\label{sec:introduction}

Many high-level computer vision tasks like object tracking, activity recognition and camera surveillance rely on a pixel-level segmentation of scenes into foreground and background as a preprocessing step. This task is often referred to as background subtraction and has drawn great attention in recent years. Surveying the multitude of existing methods is beyond the scope of this article, and for this purpose we refer to two excellent recent surveys of the field, \cite{Shireen2008} and \cite{Hassanpour2011}.

Robust Principal Component Analysis algorithms have been proven successful at separating foreground objects from a static or dynamic background \cite{Guyon2012}. The underlying assumption of Robust PCA is that the analyzed data can be considered a superposition of a low-rank and a sparse component, which can be separated blindly without any further assumptions on the data. For many video sequences this assumption holds true. The vectorized frames of a video background span a low-dimensional subspace, whereas rapidly moving objects appear sparse in space and time and thus can be distinguished from the background using Robust PCA.
Most Robust PCA algorithms focus on processing the complete data set at once in a batch-processing manner. This means that all frames of the video sequence and their statistics are available and the algorithm performs background subtraction on the entire sequence. Recently, methods have been presented which allow for online subspace tracking \cite{Balzano2010}, i.e.~video data can be processed frame by frame and each new incoming data sample contributes to the estimate of the background.

This paper introduces a robust online background subtraction algorithm, called \textit{pROST}: a smoothed $\ell_{\underline{p}}$-norm \underline{R}obust \underline{O}nline \underline{S}ubspace \underline{T}racking Method. The name reflects the two defining characteristics of the algorithm. Firstly, to achieve robustness against outliers we use a smoothed and weighted $\ell_p$-(pseudo)-norm cost function. Secondly, an efficient alternating online optimization framework for the estimating the subspace makes the algorithm suitable for online subspace tracking. The algorithm is tailored for real-time background subtraction in streaming video and makes use of the spatio-temporal dependencies between pixel labels, i.e.~the foreground or background assignment on a pixel level. This leads to especially good performance in videos that require bootstrapping, which means learning a new background from corrupted data. It also alleviates problems with large foreground objects, which often arise in PCA-based methods \cite{Bouwmans2009}. In comparison to other methods we observe that our algorithm is particularly good at dealing with the varying background in video recorded by jittery cameras.

One of the main difficulties with comparing different background subtraction methods has been the lack of an accepted benchmark. Various data sets exist (e.g.~\cite{Li2004} and \cite{Toyama1999}), which provide video sequences and manually segmented test images. However, the lack of pixel-level ground truth has led to rather selective evaluation instead of comparable and representative results, as the authors of the \emph{SABS} dataset criticize \cite{Brutzer2011}. They overcome the cumbersome task of hand-segmenting video sequences by providing an artificially rendered scene, which allows a very detailed and precise segmentation. Although the animations are close to photo-realistic, the visual impression is fundamentally different from true recordings.

In order to establish a benchmark on real-world video sequences, the \emph{changedetection.net} dataset \cite{Goyette2012} has been introduced. It provides image sequences and full ground truth for a variety of categories such as static and dynamic background, thermal imaging and camera jitter, as well as the explicit distinction between foreground objects and their shadows. Seven statistical error measures are computed to evaluate the performance as detailed as possible. This prohibits tuning a method for a single performance measure and guarantees significant scores. Evaluation and thus the ranking of all competing methods are computed per category and as an overall average. All reported results are conveniently accessible on the project website \footnote{www.changedetection.net}.

The paper is outlined as follows: in Section \ref{sec:rpcabgmodels} of this paper we define our understanding of foreground and background, give a brief overview of the issues arising in background learning and maintenance and explain how a background model can be used for foreground segmentation. We describe how PCA can be used to create a model of the scene background and motivate the use of robust cost functions. In Section \ref{sec:algorithm} we present the pROST framework, which is motivated and discussed in the context of background modeling and foreground segmentation. Section \ref{sec:implementation} provides details on the implementation on a graphics processing unit. We evaluate our algorithm on the \emph{changedetection.net} dataset and discuss the results in Section \ref{sec:evaluation}. Concluding the paper, we analyze typical issues in the modeling of scene backgrounds with pROST and explain with a few examples how they are addressed by the choice of parameters.

\section{Robust PCA based background models}
\label{sec:rpcabgmodels}

Video background is commonly defined as the union of persistent elements of a scene. They can be static or may exhibit repetitive dynamics, which either occur on an object-level, e.g.~an escalator or a fountain, or on a global scale, e.g.~water or waving trees, but also camera jitter. In other words, the background is comprised of elements that are known, predictable and not of interest for higher-level tasks such as surveillance or activity recognition. Everything else that moves about in the scene is understood as foreground objects. From this definition the idea of treating foreground-background segmentation as an outlier detection problem arises naturally, i.e.~a model of the background is established and the foreground is segmented by comparing each video frame to this model. Elements of the video frame that do not fit the background model are labeled as foreground, while the rest is labeled as background. Virtually every algorithm published so far follows this approach, but they differ in the type of background model that is used and how it is maintained, and in potential pre- and post-processing steps.

Establishing and maintaining an accurate background model is not trivial under real-world circumstances, and certain requirements have to be met by a method to be useful in challenging scenarios. For example, a scene cannot be observed in all possible lighting or weather conditions. Or it might be impossible to have a separate training stage in which the scene is free of foreground objects. Therefore, the ability to learn a background model from corrupted training data is of crucial importance. Without having any pixel semantics this is of course only possible if foreground objects have different statistical properties in time and space than the background. Furthermore, it is also necessary to update the background model continuously, which is commonly referred to as \textit{model maintenance}.

\subsection{PCA background models}
\label{sec:pca_backgrounds}

PCA background models were introduced in \cite{Oliver2000} as part of a system for human activity recognition. The underlying assumption of PCA models is that a vectorized video background $B \in \mathbb{R}^{m \times n}$ can be represented as a product of a subspace-defining matrix 
\begin{align}
\label{eq:stiefeldef}
&& U \in \St_{k,m} := \{U \in \mathbb{R}^{m \times k}|U^\top U = I_k\}
\end{align}
and the subspace coordinates $Y \in \mathbb{R}^{k \times n}$. $\St_{k,m}$ denotes the Stiefel manifold and $I_k$ is the $k$-dimensional identity matrix. With the supplement of a Gaussian noise matrix $E \in \mathbb{R}^{m \times n}$ whose $(i,j)$-entries $\epsilon_{i,j} \sim \mathcal{N}(0,\sigma)$ are independent Gaussian random variables, this results in the data model
\begin{align}
	&& B  =  UY + E .
\end{align}
 
Considering now a video sequence $X \in \mathbb{R}^{m\times n} $, which might also contain additional foreground objects, one can try to recover the matrix $U$ that defines the subspace and $Y$ as the solution of the optimization problem
\begin{align}\label{eq:definition_cost}
	&& \min_{U \in \St_{k,m},Y \in \mathbb{R}^{k \times n}} \| X - UY \|_{2}
\end{align}
where $\|\cdot\|_{2}$ denotes the $\ell_2$-norm. This is equivalent to the classic PCA problem \cite{Pearson1901}, whose well-known closed-form minimizer is given by the leading $k$ eigenvectors of the data covariance matrix. 

Given a basis $U$ for the background subspace and an observed image $x \in \mathbb{R}^m$ a foreground segmentation mask $F\in \{0,1\}^m$ can be obtained through fitting the model by firstly solving
\begin{align}
 && y^* = \arg \min_{y\in \mathbb{R}^k}\| x - U y\|_2
\end{align}
followed by applying a thresholding operation
\begin{align}
	&& F_{i} = \left\{\begin{array}{cl} 1 & \:\: \mbox{if }|x_i-u^\top_i y^*|>\delta\\ 0 & \:\: \mbox{else}, \end{array}\right . \quad 
\end{align}
where $u^\top_i$ denotes the $i$-th row of $U$ and $\delta > 0$ is a threshold parameter.

PCA-based methods for background estimation show great advantages over competing methods whenever backgrounds are dynamic and include illumination changes \cite{Bouwmans2009}. However, as the classical PCA approach uses the $\ell_2$-norm, the subspace reconstruction may severely suffer from outliers. In the practical context of background modeling, several problems can be observed: Firstly, in common PCA undue weight is given to the foreground elements when fitting the background model to camera frames during the segmentation process, which severely limits the admittable size of foreground objects. Secondly, images containing foreground objects can lead to corruption of the background model during bootstrapping and background maintenance. Finally, batch-processing results in tremendous memory requirements.

As discussed in \cite{Bouwmans2009}, a lot of effort has been spent to overcome these limitations. The predominant mechanism for achieving robustness in PCA-based methods is weighting or replacing individual pixels in order to reduce the influence of known foreground objects on the background reconstruction error. Adaptive thresholding \cite{Xu2006} has been proposed to allow for larger foreground objects and an attempt at a robust incremental estimation of backgrounds has been presented in \cite{Li2004b}. 

\subsection{Robust background estimation}

The computer vision community has recognized that Robust PCA methods offer substantial advantages over classic PCA and background modeling has become increasingly popular as an application of Robust PCA algorithms. For an overview of background subtraction using Robust PCA we refer to \cite{Guyon2012}. 

The typical assumption for Robust PCA \cite{Candes2011} is a data model
\begin{align}
	&& X = L + S,
\end{align}
where $S \in \mathbb{R}^{m\times n}$ is sparse (i.e.~having few non-zero entries) and $L \in \mathbb{R}^{m\times n}$ is low-rank. Under mild assumptions on $L$ and $S$ it is possible to recover them via
\begin{align}
\label{eq:idealrpca}
	&& (L^\ast, S^\ast) = \Argmin{\rk{L} \leq k} \|S\|_0, \; \mbox{s.t.} \; X = L+S .
\end{align}

A Robust PCA method is proposed in \cite{Candes2011} that performs a convex relaxation of \eqref{eq:idealrpca} employing an $\ell_1$-penalized outlier matrix $S$ and minimization of $\|L\|_\ast$, which denotes the nuclear norm. Under specific circumstances this convex method is able to recover the low-rank component exactly. However, only a whole batch of data samples can be processed and the proposed solvers do not achieve realtime performance. The authors of \emph{GoDec} \cite{Zhou2011} report a significantly faster processing time, which is achieved by using random projections. The method is robust against additive Gaussian noise, but it requires an estimate on the cardinality of the sparse component.

A different way of searching for a low-rank approximation of given data is to employ the so-called Grassmannian, which is the manifold of fixed-dimensional subspaces \cite{Boumal2011}, \cite{Keshavan2010}. In \cite{Balzano2010} it is shown how the Grassmannian can be exploited for online subspace tracking, i.e.~analyzing data sample-wise and constantly adapting its low-rank approximation using a one-step gradient descent. The authors furthermore demonstrate that subspaces can be reconstructed even from highly subsampled data if the upper bound on the desired rank is very low compared to the dimension of the data, which is in the same spirit as \cite{Waters2011}. Finally, the \emph{GRASTA} method \cite{He2012} robustifies subspace tracking using an $\ell_1$-norm cost function and achieves close to realtime performance on an online background subtraction task.

Approximating the impractical $\ell_0$-norm by the $\ell_1$-norm offers the advantage of obtaining a convex problem with a guaranteed globally optimal solution. However, it is known that other measures such as an $\ell_p$-norm offer a better approximation of the $\ell_0$-norm, cf.~\cite{Gasso2009}. In \cite{Hage2012} a way is shown how these kind of $\ell_0$-surrogates can be incorporated in an alternating minimization framework for robust subspace estimation and tracking. Numerical results and online background subtraction experiments indicate that using a smoothed $\ell_p$-norm sparsifying function increases the robustness of such kind of methods even further. This paper builds on the results of both \cite{Hage2012} and \cite{He2012}, and we present a realtime robust online subspace tracking method based on alternating minimization of a smoothed $\ell_p$-norm sparsifying function on manifolds using one-step gradient and conjugate gradient descent.

\section{The pROST algorithm}
\label{sec:algorithm}

As with the classic PCA based models in Section \ref{sec:pca_backgrounds}, we assume that an image $x \in \mathbb{R}^m$ is generated by a background subspace model with the addition of Gaussian noise $\epsilon \in \mathbb{R}^m$ and a sparse outlier vector $s \in \mathbb{R}^m$  which represents the foreground in the scene, i.e.
\begin{align}
	&& x  =  Uy + s + \epsilon .
\end{align}
Our goals are (i) to robustly recover this background subspace $U$ from training data containing foreground objects (i.e. $s \neq 0$), (ii) to robustly fit the model to unknown video frames in order to determine the foreground $s$, and (iii) to track any changes to the background subspace of a scene. 

\subsection{Weighted smoothed $\ell_p$-norm cost function}
\label{sec:cost}
In \cite{Hage2012} it has been shown that smoothed non-convex sparsity measures allow the reconstruction of subspaces from corrupted data in cases where other methods fail. Thus, we construct the cost function based on the smoothed $\ell_p$-norm as
\begin{align}
\label{eq:smoothlpnorm}
 && h_\mu: \mathbb{R}^{m} \rightarrow \mathbb{R}^+, x \mapsto  \sum_{i=1}^{m} \left( x_{i}^2 + \mu\right)^{\frac{p}{2}}, \:\: 0<p<1,
\end{align}
where $\mu > 0$ is a smoothing parameter. This particular $\ell_0$-surrogate serves as an arbitrary example, for other possible functions see \cite{Hage2012}. Even though using \eqref{eq:smoothlpnorm} leads to a non-convex optimization problem, in practice it is good-natured and can be optimized locally by standard methods.

The pROST algorithm is designed with background subtraction for video streams in mind, and thus we can further tailor the cost function to this setting. In video data it is sensible to assume that spatial and temporal proximity of pixels entail identical semantics. In other words, corresponding pixels in consecutive frames are likely to have the same label. This knowledge can be used to further increase the robustness of the residual cost. The idea is to reduce the contribution of labeled foreground pixels to the overall penalty by introducing additional pixel weights $w_{i} \in \mathbb{R^{+}}$, whose magnitudes depend on the labels assigned to the pixels in the previous frame. If the pixel was previously labeled a foreground pixel and is therefore likely to remain an outlier in the current frame, the weighting should be small to avoid foreground objects compromising the background. In the reverse case, if the pixel was labeled a background pixel before the weight should be equal to one to allow for model maintenance. In this way the algorithm avoids erroneously fitting the background model to already known foreground objects and it can focus on fitting the background model to the scene background instead. This extension to the cost function does not only ease bootstrapping from corrupted training data, but it also overcomes the reported difficulties of PCA methods with large foreground objects \cite{Bouwmans2009}.

We incorporate pixel-weighting by defining the weighted smoothed $\ell_p$-norm cost function
\begin{align}
  && h_{\mu,w}: \mathbb{R}^{m} \rightarrow \mathbb{R}^+, x \mapsto  \sum_{i=1}^{m} w_{i} \left( x_{i}^2 + \mu\right)^{\frac{p}{2}}, \:\: 0<p<1 \label{eq:weightedLPnorm}
\end{align}
and the eventual cost function to be minimized in pROST is
\begin{align}
	&& h_{\mu,w}( x - Uy ) \label{eq:spcacost}.
\end{align}

\subsection{Optimization on the Grassmannian}
The topics of optimization on the Stiefel manifold and the Grassmannian are covered in great detail in \cite{Absil2008} and \cite{Edelman1998}. Here we only recall the most important results and apply them to our specific problem.

In Section \ref{sec:pca_backgrounds} we define $U$ to be an element of the so-called Stiefel manifold $\St_{k,m}$. 
However, optimizing over the entire set $\St_{k,m}$ is not necessary, because
whenever $(U,y)$ is a reasonable solution then so is $(UQ,Q^\top y)$ for 
\begin{align}
	&& Q \in O(k):=\{ Q \in \mathbb{R}^{k\times k}~|~Q^\top Q = I_k \},
\end{align}
where $O(k)$ denotes the set of $k \times k$-dimensional orthogonal matrices. In other words, we are only interested in the subspace spanned by the columns of $U$, and not in a particular basis of that subspace, so the search space can be reduced. To that end we employ the well known Grassmannian, which is defined as the quotient manifold
\begin{align}
	&& \Gr_{k,m}:=\St_{k,m}/O(k),
\end{align} 
with the equivalence relation $U \sim \tilde{U}$ if and only if there exists a $Q \in O(k)$ such that $\tilde{U}=UQ$. We denote the equivalence class for some representative $U \in \St_{k,m}$ by
\begin{align}
	&& [U]=\{\tilde{U} \in \St_{k,m}~|~\tilde{U} \sim U \} \in \Gr_{k,m}.
\end{align}
Note that the class $[U]$ does not have a matrix representation in $\mathbb{R}^{m \times k}$. So whenever we store $[U]$, we will do that by using one (arbitrary) class representative. 
In contrast, the Stiefel manifold has a unique matrix representation, as does its tangent space, which is given by
\begin{align}
	&& T_U \St_{k,m}:=\{ B \in \mathbb{R}^{m \times k}~|~B^\top U+U^\top B =0 \}.
\end{align}
Due to the quotient geometry of $\Gr_{k,m}$, the spaces $T_U \St_{k,m}$ and $T_{UQ} \St_{k,m}$ share one subspace independently of $Q$, which we identify with the tangent space of the Grassmannian, namely
\begin{align}
	&& T_{[U]} \Gr_{k,m}:=\{ B \in \mathbb{R}^{m \times k}~|~U^\top B =0 \}.
\end{align}

Optimization on Riemannian manifolds like the Grassmannian is done by moving along tangent space directions on geodesics of the manifold. In order to do this, it is necessary to first project the ambient space gradient \begin{align}
&& \mathfrak{grad}_U = \nabla_U h_{\mu,w}( x - Uy )
\end{align}
onto the tangent space at $U$ and then, in the case of minimization, move along the geodesic on the manifold in the opposite direction. 

Geodesics are curves that locally minimize the distance between two points on the manifold. For a given tangent direction $B \in T_{[U]} \Gr_{k,m}$, the geodesics on the Grassmannian emanating from $[U]$ in direction $B$ are given by $[U(t)]$ with 
\begin{align}
	&& U(t)=(U V \cos(\Sigma t) + \Theta \sin(\Sigma t))V^\top, \label{eq:geodesic}
\end{align}
where $\Theta\Sigma V^\top=B$ is the compact Singular Value Decomposition of $B$, cf.~\cite{Edelman1998}.

It is easy to verify that the orthogonal projection of some $H \in \mathbb{R}^{m \times n}$ onto $T_{[U]} \Gr_{k,m}$ is given by
\begin{align}
	&& \pi(H)=(I - U U^\top)H \label{eq:gproject}.
\end{align}

Using all these components we can formulate a procedure in order to find the background subspace model $U$. We propose an alternating approach that iteratively updates $U$ and $y$. The cost function that is minimized is invariant on equivalence classes $[U]$ when considered with two variables $(U,y)$. However, as this is no longer the case when $y$ is fixed, it is not reasonable to search for an optimal element on $\Gr_{k,m}$. Thus, the optimization step for fixed $y$ will be taken in the direction $\pi(\mathfrak{grad}_U)$ along $U(t)$ as defined in \eqref{eq:geodesic}.

\subsection{An online alternating minimization algorithm}

In an online setting the video frames arrive at a certain rate and have to be processed as they arrive. Processing a frame $x^{(i+1)} \in \R^m$ at time instance $i+1$ involves three steps, which are robustly fitting the background model to the frame, updating the background subspace model $U$ to cope with changes in the background, and segmenting foreground and background for the current frame:

\paragraph{Step 1:}
Refine $y^{(i)}$ to obtain $y^{(i+1)}$ via 
\begin{align}
	&& y^{(i+1)} = \arg \min_{y \in \mathbb{R}^{k}} h_{\mu,w}( x^{(i+1)} - U^{(i)}y ) \label{eq:onlinestep1}
\end{align}

\paragraph{Step 2:}
Take one gradient descent step along $U(t)$ as defined in \eqref{eq:geodesic} to approximate
\begin{align}
 &&  t^\ast = \Argmin{t \in \mathbb{R}}\ h_{\mu,w}(x^{(i+1)} - U(t)y^{(i+1)} )
\end{align}
and to obtain the updated subspace $U^{(i+1)}:=U(t^\ast)$.

\paragraph{Step 3:}
Identify the outliers or the foreground pixels to obtain the reconstruction cost weighting $w$ for the next iteration
\begin{align}
\label{eq:pixelweighting}
	&& w_{i} = \left\{\begin{array}{cl} 1 & \:\: \mbox{if }|x_i-u^\top_i y|<\delta\\ \omega & \:\: \mbox{else}, \end{array}\right . \quad 
\end{align}
where $\omega$ is the weight for the foreground pixel reconstruction error. In order to be able to slowly incorporate foreground objects into the background, $\omega$ should be set to a small, but non-zero value.

In Step 1 pROST uses a Conjugate Gradient (CG) algorithm \cite{Hestenes1952} to perform the optimization. Even though one iteration of CG is more expensive than one iteration of a simpler gradient descent algorithm it needs fewer iterations to identify the outliers. Since most of the computational cost per iteration is due to the evaluation of the cost function and computation of the gradient, this actually leads to a more efficient algorithm. Our experiments have shown, that in most cases as little as five CG iterations are sufficient.

In Step 2 pROST takes one gradient decent step on the Grassmannian. This would usually require the costly computation of the full SVD of the projected gradient. In the online setting, however, this can be avoided. The derivative of the cost function with respect to $U_{i,j}$ is given by
\begin{align}
	&&\frac{\partial h_{\mu,w} (r)}{\partial U_{i,j}} = -p\ w_{i}\ r_i\ ( r_i^2 + \mu)^{\left(\frac{p}{2}-1\right)}y_j
\end{align}
with $r = x - Uy$. Using the short notation 
\begin{align}
&& \eta_{i}= -p\ w_{i}\  r_i\ (r_i^2 + \mu)^{\left(\frac{p}{2}-1\right)}, \quad i=1,\dots, m
\end{align}
the projected gradient can be expressed as
\begin{align}
	&& G_U = \pi(\eta) y^\top .
\end{align}
It can easily be verified that $G_U$ has rank one and its SVD is given by
\begin{align}
&&G_U &= \sigma_1 s v^\top \quad \mbox{with}\\
&& s &= \frac{\pi(\eta)}{\|\pi(\eta)\|_2},\quad \sigma_1 = \|\pi(\eta)\|_2 \|y\|_2,\quad	v = \frac{y}{\|y\|_2}. \label{eq:svd}
\end{align}
Consequently, we are freed of computing the SVD of the search direction at each iteration. This approach has also been taken in GROUSE \cite{Balzano2010} and GRASTA \cite{He2012} in order to obtain a fast online gradient decent algorithm for subspace estimation.

\subsection{Practical issues}

\paragraph*{Initialization} \quad In this alternating scheme an initialization for $U$ has to be provided. We choose to initialize the subspace randomly, which can be performed by computing a reduced QR-decomposition of a random $m \times k$ matrix. This means that pROST does not use a separate batch initialization phase. It is fully capable of recovering subspaces from video data corrupted with foreground objects. The background subspace is learned one frame at a time while continually reducing the step size from $t_{init}$ to $t_{min}$. The former should be chosen quite high ($t_{init} \in [10^{-4},1]$) to facilitate quick initialization, while the latter should be chosen quite low to avoid trailing ghost images of moving foreground objects ($t_{min} \in [10^{-7},10^{-4}]$).

The step size for the subspace updates in each iteration are defined by the step-size rule
\begin{align}
	&& t = \max \{ e^{- \tau i} t_{init}, t_{min} \},
\end{align}
where $i$ is the iteration and $\tau$ is a parameter controlling the shrinkage rate for the step size reduction. Whenever an initialization phase is defined by an exact number of frames $i_{init}$, the parameter $\tau$ can be calculated as
\begin{align}
	&& \tau = \frac{-\log\left(\frac{t_{min}}{t_{init}}\right)}{i_{init}} \ .
\end{align}

\paragraph*{Pre- and post-processing} \quad
Firstly, the running average of the image data is maintained during the initialization phase and subtracted from each frame before pROST is applied. This means the background subspace has to capture only the dynamic aspects of the scene. Secondly, the images are normalized by dividing the intensity values by the sample standard deviation over all pixels in the initialization phase. In our experiments we observe that this kind of preprocessing is highly beneficial for capturing the scene dynamics. To achieve fast and uniform processing we re-sample all videos to a size of $160 \times 120$.

Apart from the thresholding operation, we also apply a $3 \times 3$ median filter to the foreground segmentation mask $F$ to fill small holes and to get rid of small clusters of erroneously labeled pixels.

\paragraph*{Color images} \quad
If colored video is available, it is clearly advantageous to use the information provided in the color channels for segmentation. We represent a colored vectorized video frame of size $m$ by a vector
\begin{align}
	&& x=\begin{bmatrix} x_R \\ x_G \\ x_B \end{bmatrix} \in \mathbb{R}^{3m},
\end{align} 
where the $i$-th entry of $x_R, x_G, x_B \in \mathbb{R}^m$ is given by the respective channel value at pixel $i$. Accordingly, the background subspace is modeled by
\begin{align}
	&& U=\begin{bmatrix} U_R \\ U_G \\ U_B \end{bmatrix} \in \St_{k,3m}.
\end{align}
The pixel $i$ is classified as foreground if the difference between the reconstructed background of either of the channels is large enough, i.e. if
 \begin{align}
	&& \max\{| x_{R,i} - u_{R,i}^\top y  |, |x_{G,i} - u_{G,i}^\top y| , |x_{B,i} - u_{B,i}^\top y|\} \geq \delta. \end{align}
Here, $u_{R,i}$, $u_{G,i}$, $u_{B,i}$ denote the respective rows of $U$.

\section{Implementation}
\label{sec:implementation}

In order to achieve realtime performance we have implemented pROST on a GPU. More precisely, the preprocessing and all steps of pROST are implemented on the GPU, whereas the median filtering operation in the post-processing stage runs on the CPU. For transferring the images to the GPU we use pinned host memory.

One of the strengths of pROST is its simplicity. Since most of the operations involved are matrix operations, pROST can be parallelized very efficiently on a GPU using C++, CUDA and the highly optimized CUBLAS library for linear algebra operations on the GPU. Step 2 of pROST, for example, can be implemented with as little as four General Matrix Multiply (GEMM) operations and only takes about 5~ms for a subspace dimension of $k=15$ and an image resolution of $160 \times 120$ on a Nvidia GTX 660 GPU.

For images in this resolution and a subspace dimension of $15$, more than $50\%$ of the computation time is spent on matrix multiplications, further $30\%$ on basic operations like matrix addition, element-wise multiplication and evaluating the cost function. In order to optimize the implementation even further, we have taken great care to reduce the required number of matrix multiplications and order them in such a way as to reduce the overall complexity and memory requirements. Evaluating the cost function involves a parallel reduction, which has been implemented following the scheme presented in \cite{Harris2008}.
\begin{figure}
\label{fig:fps_performance}
\caption{Performance of the GPU implementation of pROST for several image scalings and subspace dimensions. Scaling is relative to $320 \times 240$ images.}

\begin{tikzpicture}
    \begin{axis}[
      ylabel=Frames per second (FPS),
      xlabel=Scaling,
      legend style={at={(0.8,0.9)},anchor=north}
      ] 
        \addplot+[sharp plot,color=Diagram1,no markers,line width=2] coordinates {
            (0.125, 66) 
            (0.25, 52) 
            (0.5, 29) 
            (1.0,9) 
            };
        \addplot+[sharp plot,color=Diagram6,no markers,line width=2] coordinates {
            (0.125, 70) 
            (0.25, 66) 
            (0.5, 34) 
            (1.0,13) 
            };
        \addplot+[sharp plot,color=Diagram5,no markers,line width=2] coordinates {
            (0.125, 77) 
            (0.25, 76) 
            (0.5, 46) 
            (1.0,14) 
            };

        \legend{pROST k=15,pROST k=5, pROST k=2}
    \end{axis}
\end{tikzpicture}
\end{figure}
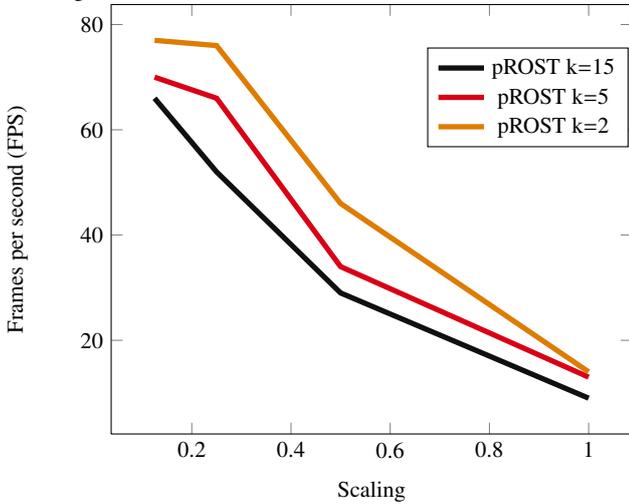

\section{Evaluation}
\label{sec:evaluation}
Our goal for the evaluation is twofold. Firstly, we rank the pROST method for background subtraction among competing methods on a widely-known benchmark. Secondly, we show that using the weighted smoothed $\ell_p$-norm instead of the $\ell_1$-norm leads to superior results for background subtraction by comparing our method to GRASTA \cite{He2012}, a state-of-the-art representative of online Robust PCA.
As mentioned in Section \ref{sec:introduction} we conduct all experiments on the \emph{changedetection.net} dataset, and apart from discussing the results here we will also publish the results on the project website to allow a quick and easy comparison with different approaches like GMM-based or non-parametric methods. As the benchmark requires a static configuration for all scenarios we fix all parameters for a first overview and discuss their particular influence afterwards in a more detailed investigation.

\subsection{Performance on the \emph{changedetection.net} benchmark}
The \emph{changedetection.net} dataset \cite{Goyette2012} consists of six categories of videos and provides ground truth for each frame. The ground truth contains information about background and foreground objects as well as their boundaries and shadows. For some of the videos, the segmentation is evaluated only for certain regions of interest (ROI) while for others the whole image is evaluated. In order to produce comparable results, an evaluation tool is provided which computes significant statistical measures for the segmented images. The evaluation starts after a certain number of frames, which can be used for initialization. However, these training samples have the same foreground-background distribution as the ones used for evaluation and can therefore contain foreground objects.

For the benchmark evaluation we select the following parameters, which maximize the overall performance
\begin{itemize}
	\item $k=15$ (subspace dimension),
	\item $\omega=5\times 10^{-5}$ (foreground weighting),
	\item $t_{init}=5 \times 10^{-3}$ (initial stepsize),
	\item $t_{min}=10^{-4}$ (online stepsize),
	\item $\delta=0.35$ (threshold),
	\item $p=0.25$, $\mu=\delta^2(1-p)$ (smoothed $\ell_p$-norm parameters).
\end{itemize}
For each frame we perform a maximum of five CG steps for the optimization of \eqref{eq:onlinestep1}.

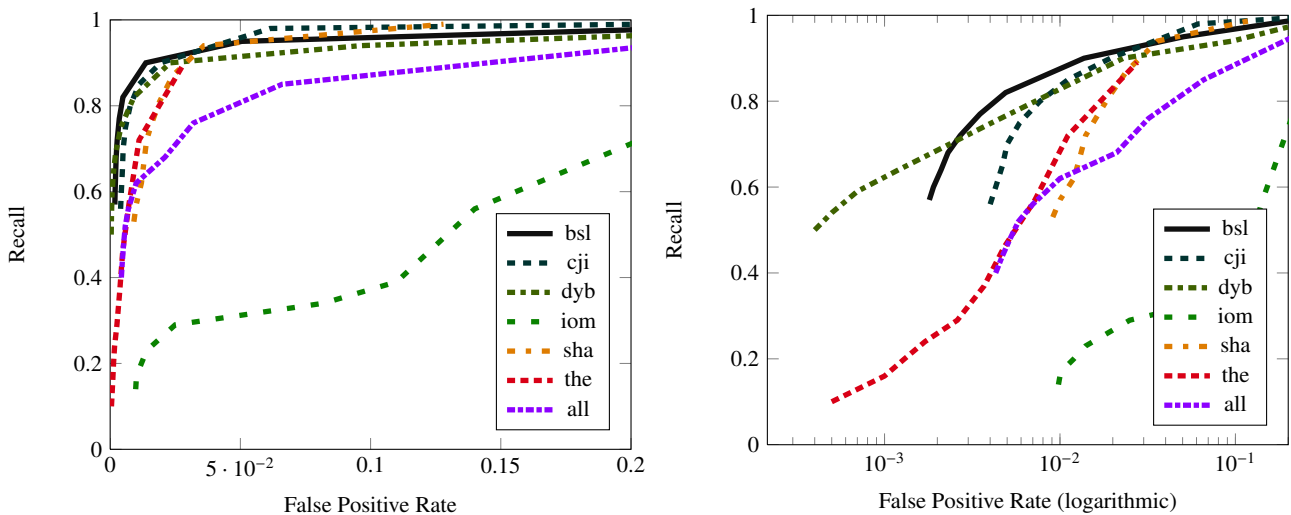
\begin{figure*}
\caption{ROC curves for pROST on the \textit{changedetection.net} benchmark (bsl: \emph{baseline}, cji: \emph{camera jitter}, dyb: \emph{dynamicBackground}, iom: \emph{intermittentObjectMotion} ,sha: \emph{shadow}, the: \emph{thermal}, all: \emph{overall})}
\label{fig:ROC}
\centering
\begin{tikzpicture}
    \begin{axis}[
      ylabel=Recall,
      xlabel= False Positive Rate,
      xmin=0,xmax=0.2,
      ymin=0,ymax=1,
      legend style={at={(0.85,0.55)},anchor=north}
      ] 
        \addplot+[sharp plot,color=Diagram1,no markers,line width = 2,solid] coordinates {
            (0.0018, 0.57) 
            (0.0019, 0.60) 
            (0.0021, 0.64) 
            (0.0023, 0.68) 
            (0.0027, 0.72)
            (0.0035, 0.77)
            (0.0049, 0.82)
            (0.0137, 0.90)
            (0.0505, 0.95)
            (0.3299, 1.0)
            };
        \addplot+[sharp plot,color=Diagram2,no markers,line width = 2,dashed] coordinates {
            (0.004, 0.56) 
            (0.0043, 0.60) 
            (0.0047, 0.65) 
            (0.0050, 0.70) 
            (0.0059, 0.75)
            (0.0077, 0.80)
            (0.0111, 0.85)
            (0.0194, 0.90)
            (0.0614, 0.98)
            (0.3505, 1.0)
            };
            \addplot+[sharp plot,color=Diagram3,no markers,line width = 2, dashdotted] coordinates {
            (0.0004, 0.50) 
            (0.0005, 0.54) 
            (0.0007, 0.59) 
            (0.0012, 0.64) 
            (0.0021, 0.69)
            (0.0041, 0.75)
            (0.0091, 0.82)
            (0.0230, 0.90)
            (0.0961, 0.94)
            (0.3663, 1.0)
            };
            
            \addplot+[sharp plot,color=Diagram4,no markers,line width = 2,loosely dashed] coordinates {
            (0.0098, 0.14) 
            (0.01, 0.16) 
            (0.0107, 0.18) 
            (0.012, 0.20) 
            (0.014, 0.23)
            (0.025, 0.29)
            (0.0818, 0.34)
            (0.11, 0.39)
            (0.14, 0.56)
            (0.2707, 0.89)
            };
            
            \addplot+[sharp plot,color=Diagram5,no markers,line width = 2,loosely dashdotted] coordinates {
            (0.0091, 0.53) 
            (0.01, 0.57) 
            (0.012, 0.62) 
            (0.013, 0.67) 
            (0.014, 0.72)
            (0.016, 0.76)
            (0.019, 0.81)
            (0.024, 0.87)
            (0.036, 0.94)
            (0.128, 0.99)
            };
            
             \addplot+[sharp plot,color=Diagram6,no markers,line width = 2,densely dashed] coordinates {
            (0.0005, 0.10) 
            (0.001, 0.16) 
            (0.0013, 0.20) 
            (0.0017, 0.24) 
            (0.0026, 0.29)
            (0.0037, 0.37)
            (0.0048, 0.46)
            (0.0072, 0.57)
            (0.0111, 0.72)
            (0.0275, 0.89)
            };
            
            \addplot+[sharp plot,color=Diagram7,no markers,solid,line width = 2,densely dashdotted] coordinates {
            (0.0043, 0.4) 
            (0.0047, 0.44) 
            (0.0052, 0.48) 
            (0.0058, 0.52) 
            (0.0070, 0.56)
            (0.01, 0.62)
            (0.021, 0.68)
            (0.032, 0.76)
            (0.066, 0.85)
            (0.24, 0.96)
            };
            
        \legend{bsl, cji, dyb, iom ,sha, the, all}
    \end{axis}
    
\end{tikzpicture}
\begin{tikzpicture}
    \begin{axis}[
      xmode=log,
      ylabel=Recall,
      xlabel= False Positive Rate (logarithmic),
      xmin=0,xmax=0.2,
      ymin=0,ymax=1,
      legend style={at={(0.85,0.55)},anchor=north}
      ] 
        \addplot+[sharp plot,color=Diagram1,no markers,line width = 2,solid] coordinates {
            (0.0018, 0.57) 
            (0.0019, 0.60) 
            (0.0021, 0.64) 
            (0.0023, 0.68) 
            (0.0027, 0.72)
            (0.0035, 0.77)
            (0.0049, 0.82)
            (0.0137, 0.90)
            (0.0505, 0.95)
            (0.3299, 1.0)
            };
        \addplot+[sharp plot,color=Diagram2,no markers,line width = 2,dashed] coordinates {
            (0.004, 0.56) 
            (0.0043, 0.60) 
            (0.0047, 0.65) 
            (0.0050, 0.70) 
            (0.0059, 0.75)
            (0.0077, 0.80)
            (0.0111, 0.85)
            (0.0194, 0.90)
            (0.0614, 0.98)
            (0.3505, 1.0)
            };
            \addplot+[sharp plot,color=Diagram3,no markers,line width = 2,dashdotted] coordinates {
            (0.0004, 0.50) 
            (0.0005, 0.54) 
            (0.0007, 0.59) 
            (0.0012, 0.64) 
            (0.0021, 0.69)
            (0.0041, 0.75)
            (0.0091, 0.82)
            (0.0230, 0.90)
            (0.0961, 0.94)
            (0.3663, 1.0)
            };
            
            \addplot+[sharp plot,color=Diagram4,no markers,line width = 2,loosely dashed] coordinates {
            (0.0098, 0.14) 
            (0.01, 0.16) 
            (0.0107, 0.18) 
            (0.012, 0.20) 
            (0.014, 0.23)
            (0.025, 0.29)
            (0.0818, 0.34)
            (0.11, 0.39)
            (0.14, 0.56)
            (0.2707, 0.89)
            };
            
            \addplot+[sharp plot,color=Diagram5,no markers,line width = 2,loosely dashdotted] coordinates {
            (0.0091, 0.53) 
            (0.01, 0.57) 
            (0.012, 0.62) 
            (0.013, 0.67) 
            (0.014, 0.72)
            (0.016, 0.76)
            (0.019, 0.81)
            (0.024, 0.87)
            (0.036, 0.94)
            (0.128, 0.99)
            };
            
             \addplot+[sharp plot,color=Diagram6,no markers,line width = 2,densely dashed] coordinates {
            (0.0005, 0.10) 
            (0.001, 0.16) 
            (0.0013, 0.20) 
            (0.0017, 0.24) 
            (0.0026, 0.29)
            (0.0037, 0.37)
            (0.0048, 0.46)
            (0.0072, 0.57)
            (0.0111, 0.72)
            (0.0275, 0.89)
            };
            
            \addplot+[sharp plot,color=Diagram7,no markers,solid,line width = 2,densely dashdotted] coordinates {
            (0.0043, 0.4) 
            (0.0047, 0.44) 
            (0.0052, 0.48) 
            (0.0058, 0.52) 
            (0.0070, 0.56)
            (0.01, 0.62)
            (0.021, 0.68)
            (0.032, 0.76)
            (0.066, 0.85)
            (0.24, 0.96)
            };
            
        \legend{bsl, cji, dyb, iom ,sha, the, all}
    \end{axis}
\end{tikzpicture}
\end{figure*}

The detailed results for pROST are listed in Table \ref{tab:srpcaresults}. By varying the threshold parameter $\delta \in [0.05,0.6]$ we obtain the ROC curves for all categories, which are displayed in Figure \ref{fig:ROC}. 

\begin{table*}
\caption{Per category results for pROST in the \emph{changedetection.net} benchmark}
\label{tab:srpcaresults}
\centering
\begin{tabular}{ l || c | c | c | c | c | c | c }
  Category & Recall & Specificity & FPR & FNR & PWC & Precision & FMeasure \\
baseline &    0.801 &    0.9941 &    0.0059 &    0.199 &    1.28 & 0.805 &    0.799 \\
camera jitter &    0.770 &    0.9925 &    0.0075 &    0.230     &1.56     & 0.825 &    0.792 \\
dynamic background &    0.743 &    0.9945 &    0.0055 &    0.257 &    0.73 &    0.566 &    0.595 \\
intermittent object motion &    0.540 &    0.9137 &    0.0863 &    0.460 &    9.71 &    0.488 &    0.419 \\
shadow &    0.754 &    0.9798 &    0.0202 &    0.256 &    2.85 &    0.671 &    0.706 \\
thermal &    0.497 &    0.9920 &    0.0080 &    0.503 &    2.97 &    0.756 &    0.584 \\
overall &     0.684 &    0.9778 &    0.0222 &    0.316 &    3.18 &    0.685 &    0.650
\end{tabular}
\end{table*}

\begin{table*}
\caption{Per category results for GRASTA in the \emph{changedetection.net} benchmark}
\label{tab:grastaresults}
\centering
\begin{tabular}{ l || c | c | c | c | c | c | c }
  Category & Recall & Specificity & FPR & FNR & PWC & Precision & FMeasure \\
baseline &    0.609 & 0.9926 &    0.0074 &    0.391 &    2.13 &    0.740 &    0.664 \\
camera jitter &    0.622 &    0.9282 &    0.0718 &    0.378 &    8.36 &    0.354 &    0.434 \\
dynamic background &    0.701 &    0.9760 &    0.0240 &    0.299 &    2.61 &    0.262 &    0.355 \\
intermittent object motion &    0.311 &    0.9842 &    0.0158 &    0.689 &    6.32 &    0.515 &    0.359 \\
shadow &     0.608 &    0.9554 &    0.0446 &    0.392 &    6.09 &    0.536 &    0.529 \\
thermal &    0.344 &    0.9851 &    0.0149 &  0.656   &    6.13 &    0.726 &    0.428 \\
overall &     0.533 &    0.9702 &    0.0298 &    0.467 &    5.27 &    0.522 &    0.461
\end{tabular}
\end{table*}

In order to compare pROST to GRASTA \cite{He2012} we rely on the \textit{streaming} version of GRASTA whose MATLAB implementation is available for download on the author's website\footnote{https://sites.google.com/site/hejunzz/grasta}. This implementation is intended to work with gray scale images, whereas we work with RGB color images. To allow for a fair comparison we have modified GRASTA to work with such images. We use the same subspace dimension as for pROST and down-sample all images to a resolution of $160x120$.
GRASTA requires an initialization phase in which an initial background model is learned from a batch of training images. To allow the best possible outcome in this phase we use the largest possible set of training images, i.e.~all frames at the beginning of the videos that are not evaluated, and use all the pixels in each video frame to learn the subspace. We allow GRASTA to take three passes over the data, which means that it encounters each video frame three times as often as pROST during the initialization process. We rely on the default parameters of the MATLAB implementation except for the detection threshold and the percentage of pixels used for updating the subspace during the tracking stage. The demo implementation suggests to use 10\% of the pixels, while we use 25\% of the pixels. The reason for not using all available pixels is that GRASTA is explicitly designed for reconstructing subspaces from incomplete information. In all experiments we have observed that indeed, GRASTA's performance does not deteriorate markedly if the data is subsampled as the authors describe. We use a value of 0.2 for thresholding in GRASTA, which is twice as high as the threshold suggested by the authors. This choice is motivated by the fact, that the backgrounds in the \textit{changedetection.net} benchmark are highly dynamic and a lower threshold would lead to excessive amounts of false positives. The obtained segmentation masks are post-processed by applying a $3 \times 3$ median filter. All benchmark results for GRASTA are listed in Table~\ref{tab:grastaresults}.

\subsection{Discussion}

The pROST method excels when the camera is jittering and ranks first in this category by a large margin. In the other categories the method ranks mid-field. It is important to note, however, that it is possible to achieve better performance in the other categories by tuning the parameters individually (see Section \ref{sec:parameters}).

A strength of pROST is clearly dealing with fast variations in the background like camera jitter and scenes with quickly moving foreground objects. The outstanding performance achieved in the \textit{camera jitter} category, which mostly requires the initialization from video heavily corrupted with outliers, shows that the method can bootstrap in very difficult situations. pROST is also capable of dealing with gradual lighting changes. The comparison to GRASTA shows that pROST's performance in the \emph{camera jitter} category is not a general feature of PCA models, but rather a combined result of the cost function, the optimization methods and the introduced foreground pixel weighting.

Situations in which the algorithm fails include the relocation of background objects, for which the performance in the \textit{intermittent object motion} category is a clear indicator. These problems can be alleviated to some degree by adjusting the parameters to control the speed of background adaptation. The underlying problem, however, is that the algorithm must adapt to some changes faster than to others. When an object starts to move that was formerly a part of the background, the newly revealed background, which will now be labeled as foreground, has to be integrated into the model as quickly as possible. At the same time, the moving object has to remain in the foreground, even when it stops moving and becomes stationary. When a foreground object becomes stationary it has the same spatio-temporal properties as the newly revealed background, and consequently our algorithm will treat them equally. The demands are therefore conflicting. We argue that the background subtraction algorithm presented here is not especially designed for this task, but through the individual weighting coefficients included in the cost function, pROST could solve this problem in principle. An unresolved problem remains the occurrence of camouflaging. Besides the median filtering in the post-processing step the algorithm has no means of exploiting spatial correlation of pixel labels within a frame. It has to rely solely on color or intensity differences to make segmentation decisions at pixel level, which makes it inapt to cope with this phenomenon.

\begin{figure*}
\centering
  \subfigure[Frame 1381 of \textit{traffic} video and ground truth ]{
   \includegraphics[width=0.19\textwidth]{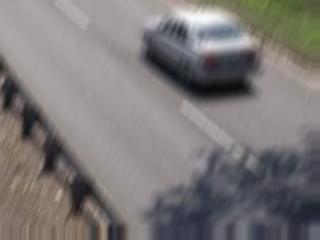}~
   \includegraphics[width=0.19\textwidth]{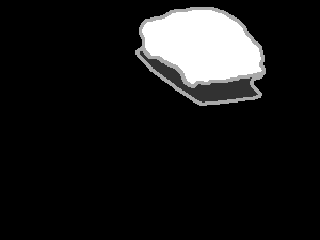}}
   \subfigure[k=5 ]{\includegraphics[width=0.19\textwidth]{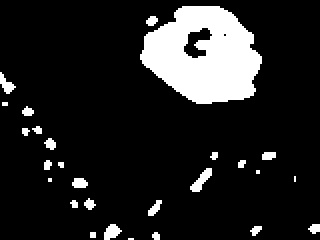}}
   \subfigure[k=15 ]{\includegraphics[width=0.19\textwidth]{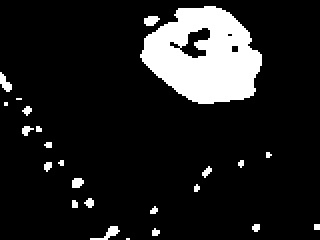}}
   \subfigure[k=40 ]{\includegraphics[width=0.19\textwidth]{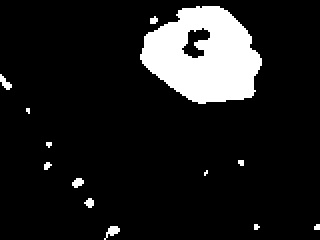}}
\caption{Effects of subspace dimensionality in jittery videos ($p=1.0$,$\mu=10^{-3},\omega=1.0$)}
\label{fig:subspace_dimensionality}     
\end{figure*}
\begin{figure*}[p]
\centering
  \subfigure[Frame 180 ]{
   \includegraphics[scale=0.35]{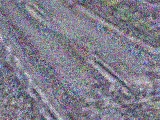}~
	\includegraphics[scale=0.35]{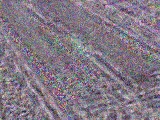}~
	\includegraphics[scale=0.35]{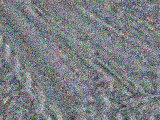}~
	\includegraphics[scale=0.35]{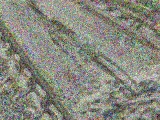}~
	\includegraphics[scale=0.35]{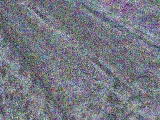}~
	\includegraphics[scale=0.35]{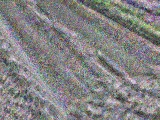}~
	\includegraphics[scale=0.35]{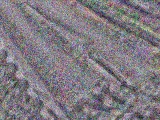}~
	\includegraphics[scale=0.35]{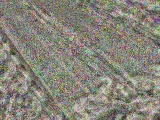}~
   }
    \subfigure[Frame 500 ]{
   \includegraphics[scale=0.35]{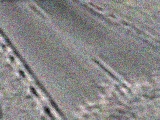}~
	\includegraphics[scale=0.35]{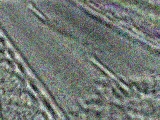}~
	\includegraphics[scale=0.35]{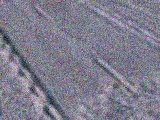}~
	\includegraphics[scale=0.35]{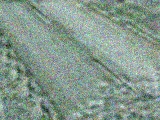}~
	\includegraphics[scale=0.35]{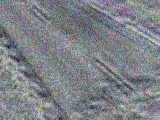}~
	\includegraphics[scale=0.35]{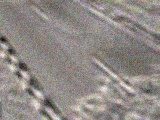}~
	\includegraphics[scale=0.35]{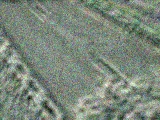}~
	\includegraphics[scale=0.35]{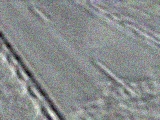}~
   }
    \subfigure[Frame 1000 ]{
   \includegraphics[scale=0.35]{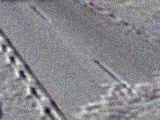}~
	\includegraphics[scale=0.35]{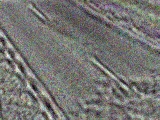}~
	\includegraphics[scale=0.35]{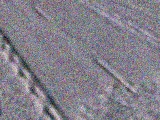}~
	\includegraphics[scale=0.35]{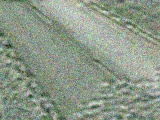}~
	\includegraphics[scale=0.35]{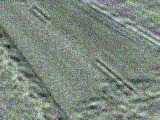}~
	\includegraphics[scale=0.35]{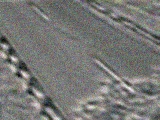}~
	\includegraphics[scale=0.35]{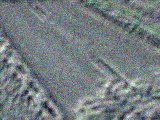}~
	\includegraphics[scale=0.35]{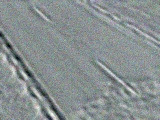}~
   }
\caption{Elements of the evolving 15 dimensional subspace for the \textit{traffic} dataset.}
\label{fig:bgmodeling_traffic}       
\end{figure*}
\begin{figure}[b]
\centering
  \subfigure[Frame 467 of \textit{bungalows} video and ground truth ]
  {\includegraphics[width=0.23\textwidth]{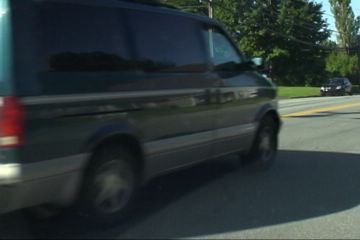}~
  \includegraphics[width=0.23\textwidth]{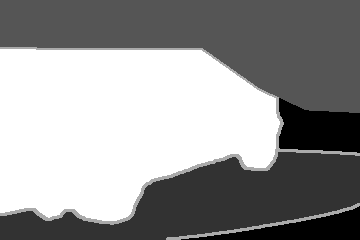}}
   \subfigure[no foreground weighting ]
   {\includegraphics[width=0.23\textwidth]{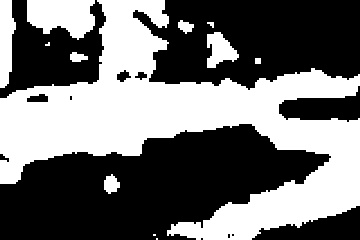}~
   \includegraphics[width=0.23\textwidth]{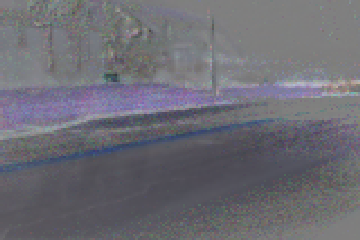}~}\\
   \subfigure[with foreground weighting $\omega=10^{-3}$ ]
   {\includegraphics[width=0.23\textwidth]{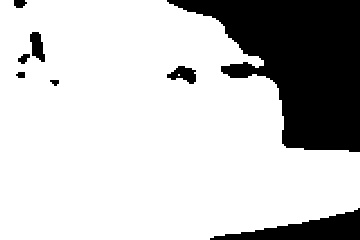}~
   \includegraphics[width=0.23\textwidth]{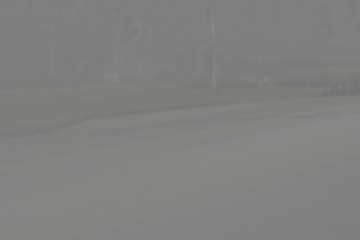}}
\caption{Effects of foreground weighting on dynamic backgrounds with extremely large foreground objects ($k=15$, $p=0.25$, $\mu=10^{-3}$)}
\label{fig:fg-weighting_bungalows}
\end{figure}
\begin{figure*}[p]
\centering
  \subfigure[Frame 2622 of the \textit{fall} video and ground truth ]{\includegraphics[width=0.23\textwidth]{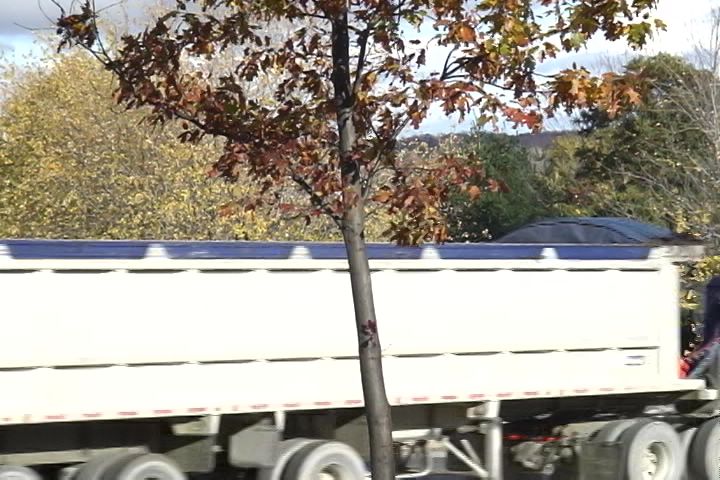}~
  \includegraphics[width=0.23\textwidth]{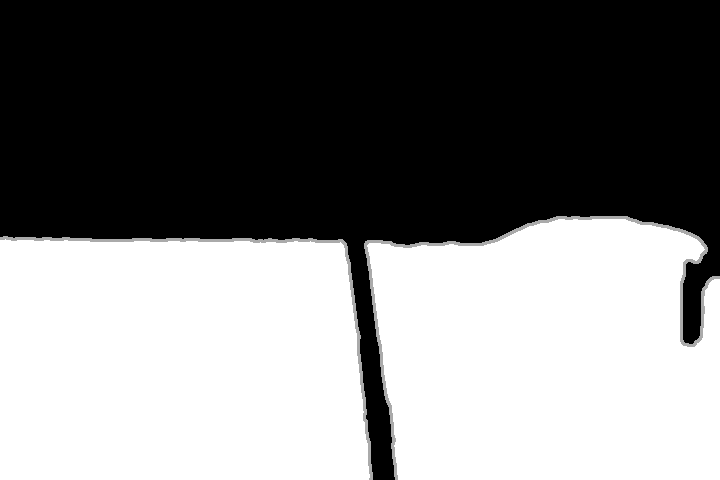}}
   \subfigure[Frame 2622, no weighting and small step size ($t_{min}=10^{-5}$)]
   {\includegraphics[width=0.23\textwidth]{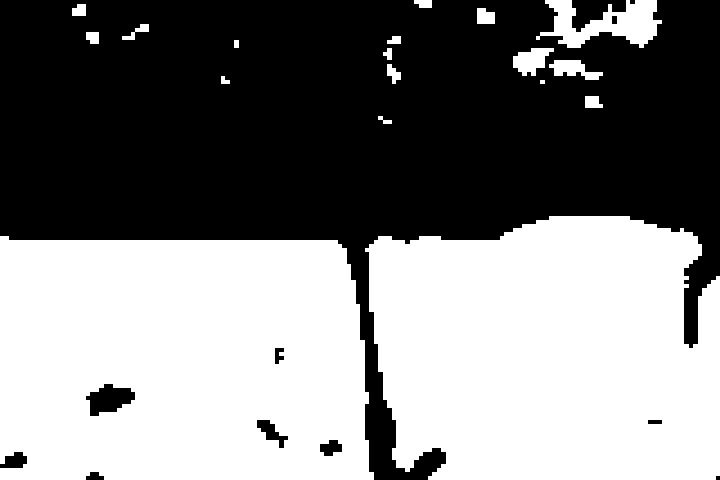}~
    \includegraphics[width=0.23\textwidth]{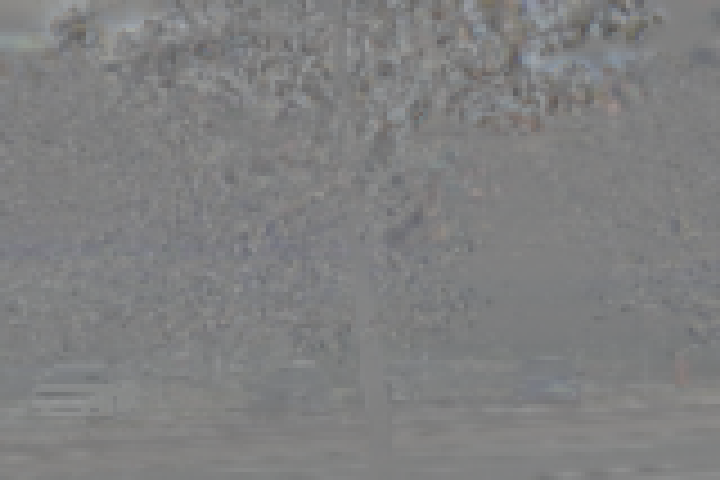}~}
   \subfigure[Frame 2049, no weighting and large step-size ($t_{min}=10^{-4}$)]{
   \includegraphics[width=0.23\textwidth]{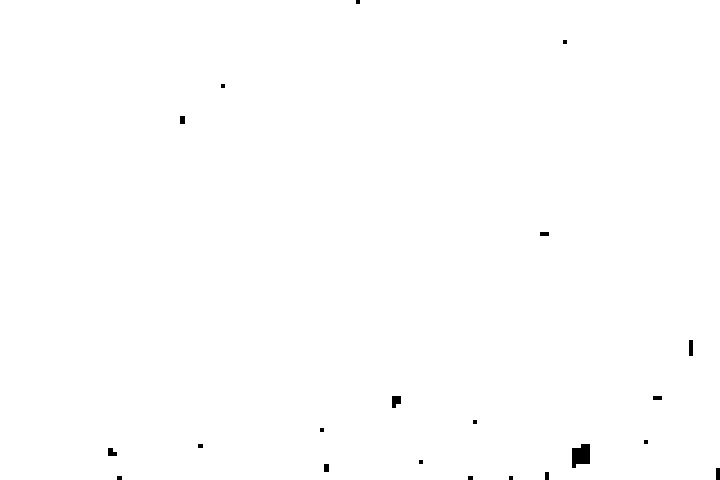}~
      \includegraphics[width=0.23\textwidth]{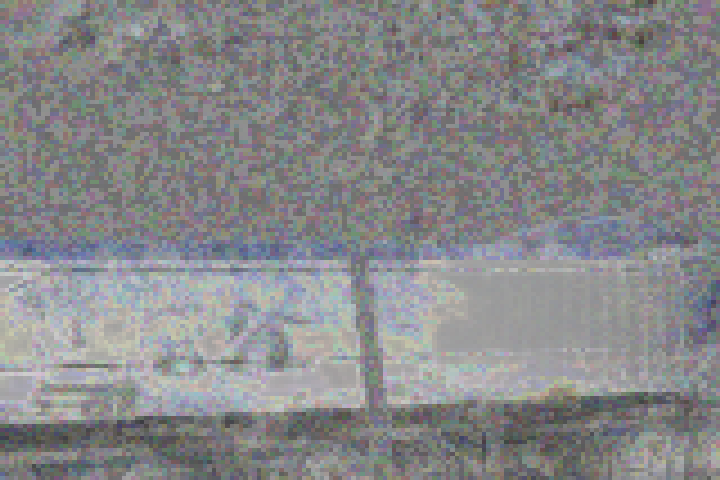}~
   }
      \subfigure[Frame 2622, weighting and large step-size ($t_{min}=10^{-4}$)]{
   \includegraphics[width=0.23\textwidth]{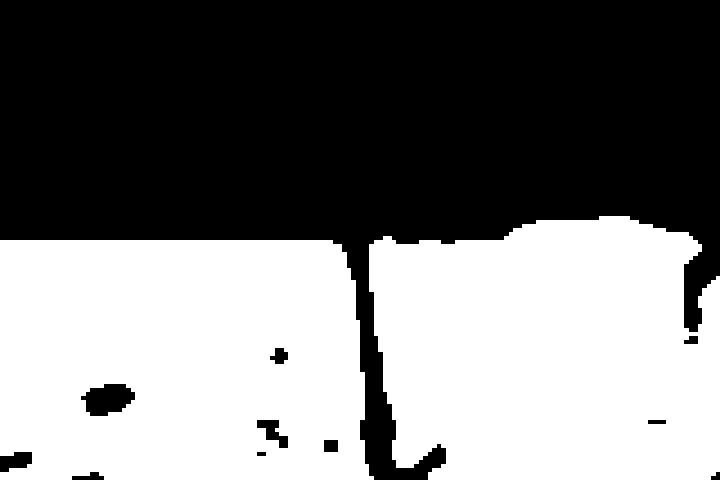}~
      \includegraphics[width=0.23\textwidth]{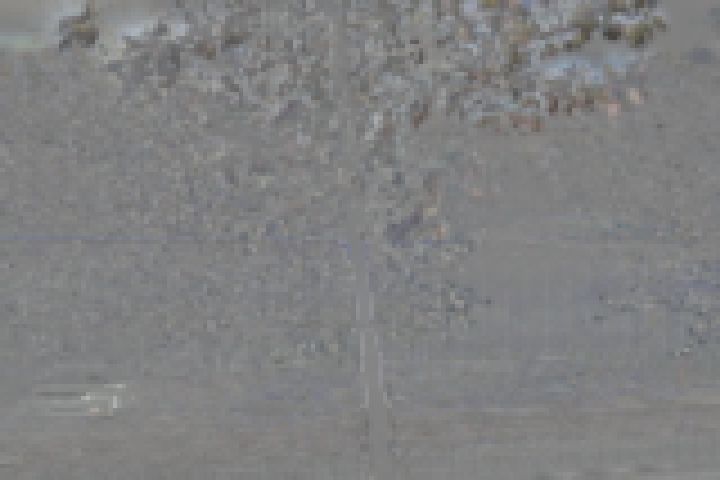}~
   }\\
\subfigure[Frame 2643 of the \textit{fall} video and ground truth ]{
   \includegraphics[width=0.23\textwidth]{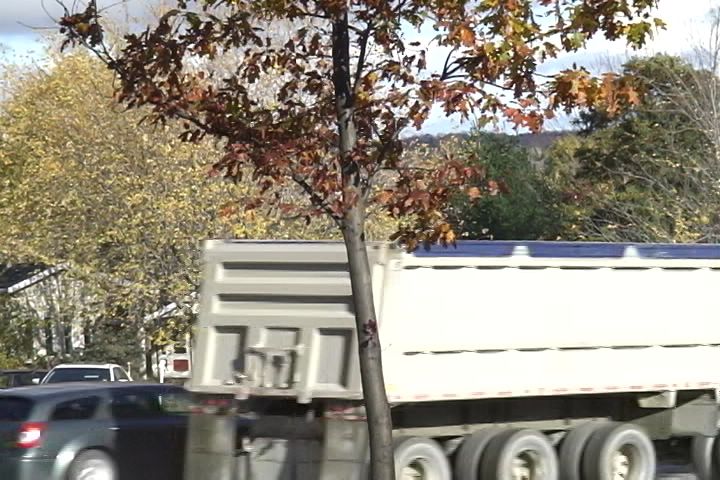}~
   \includegraphics[width=0.23\textwidth]{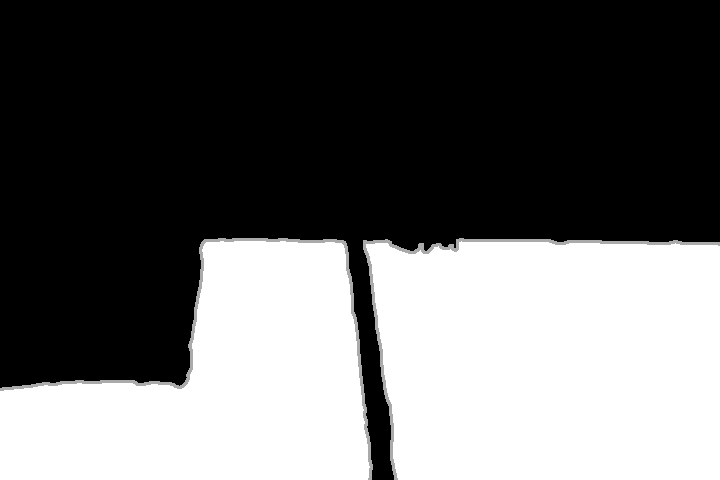}~
   }
   \subfigure[Frame 2643, no weighting and small step size ($t_{min}=10^{-5}$)]{
   \includegraphics[width=0.23\textwidth]{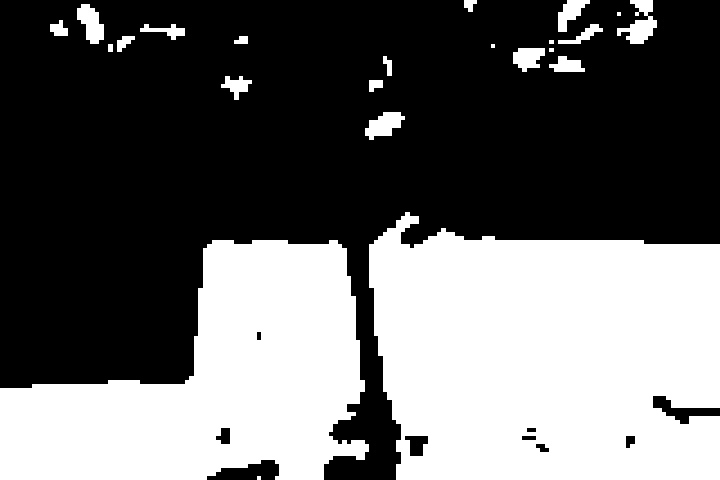}~
      \includegraphics[width=0.23\textwidth]{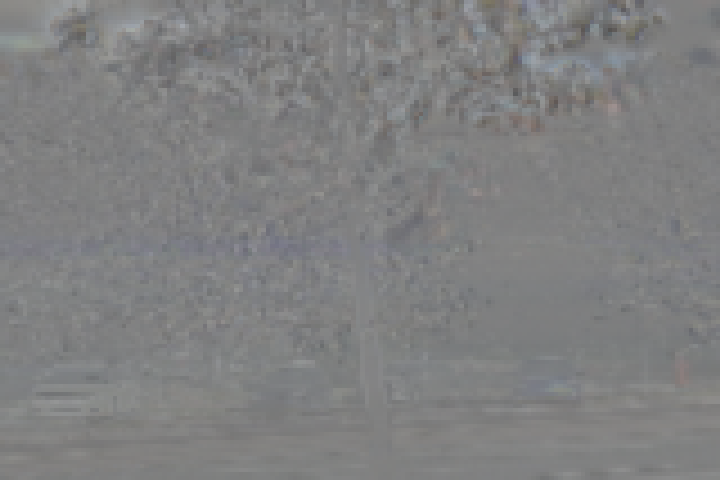}~}
   \subfigure[Frame 2643, no weighting and large step-size ($t_{min}=10^{-4}$)]{
   \includegraphics[width=0.23\textwidth]{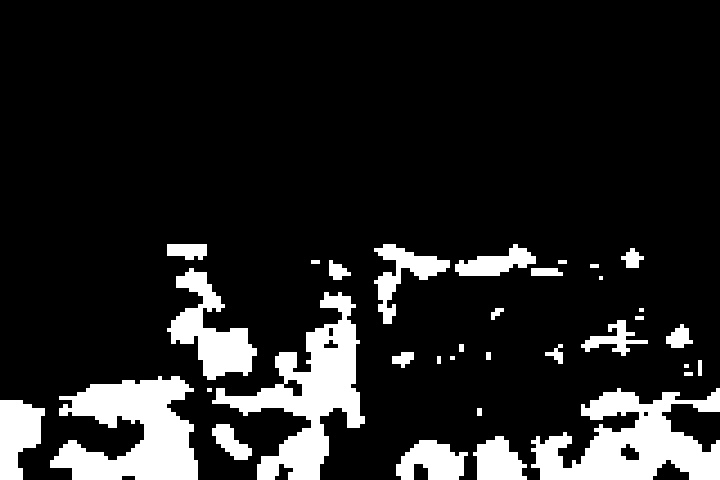}~
      \includegraphics[width=0.23\textwidth]{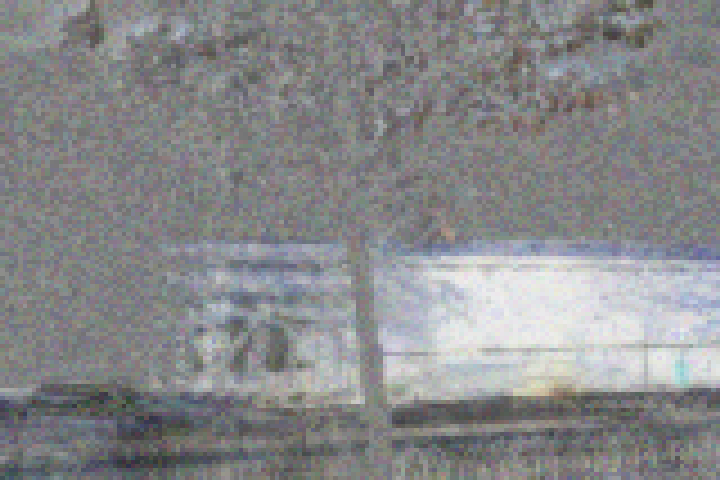}~
   }
      \subfigure[Frame 2643, weighting and large step-size ($t_{min}=10^{-4}$)]{
   \includegraphics[width=0.23\textwidth]{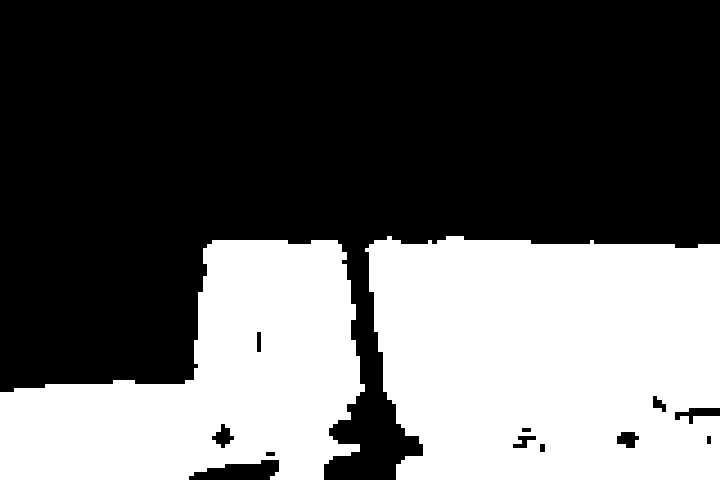}~
      \includegraphics[width=0.23\textwidth]{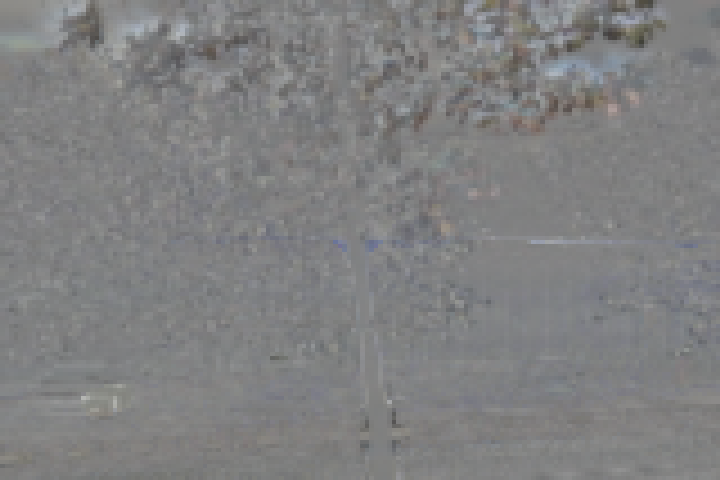}~
   }\\
\caption{Effects of foreground weighting on bootstrapping with extremely large foreground objects ($k=15$, $p=0.5$, $\mu=10^{-3}$)}
\label{fig:fg-weighting_fall}
\end{figure*}
\subsection{Choice of parameters}
\label{sec:parameters}

In the following we explain the most important parameters, their influence on the performance and behavior of pROST and how we set them for the evaluation. The effect of different parameter settings are illustrated by results obtained in the $changedetection.net$ benchmark.

\paragraph*{Subspace dimension} \quad The subspace dimension $k$ defined by the Stiefel dimension \eqref{eq:stiefeldef} can have considerable impact on the computational complexity of the algorithm, but also on other performance measures. Choosing an overly large value for $k$ leads to excessive computational complexity while not increasing performance. Low-complexity backgrounds like those in the \textit{baseline} category can be represented with a dimension as low as $k=2$, while the representation of highly dynamic backgrounds like in the \textit{camera jitter} category can benefit from a higher limit on the subspace dimension. In Figure~\ref{fig:subspace_dimensionality} the results for three different choices of $k$ and the resulting background and segmentation for frame \# $1381$ of \textit{traffic} can be seen. To get a more detailed impression of how the background is represented by pROST, we provide some further insight in Figure~\ref{fig:bgmodeling_traffic}. Notice that the background contains only the dynamic aspects of the scene as the running average is subtracted from each video frame. With growing subspace dimension finer details of the background can be captured, but the model is also getting more flexible in areas that do not require this flexibility. This causes integration of foreground objects into the background. We furthermore observed that there is a strong relationship between image sizes and required subspace dimension. For images of size $160 \times 120$ a subspace dimension of 10 to 15 is sufficient, even for the highly dynamic \textit{camera jitter} backgrounds. For images of size $320 \times 240$ the \textit{camera jitter} category requires much higher-dimensional subspaces and also a longer initialization. This problem can be mitigated by reducing the information content in the image, for example by band-limiting the image with a Gaussian blur filter. Another approach is downsampling the image before the processing and upsampling the segmentation mask for the evaluation. This is clearly preferable, because it also reduces computational complexity.

\paragraph*{Foreground weighting} \quad The foreground weighting parameter $\omega$ from \eqref{eq:pixelweighting} has a large effect on the algorithm's bootstrapping capability, how it deals with highly dynamic complex backgrounds and robustness to large foreground objects. In some scenarios like e.g.~the \textit{bungalows} video in the \textit{shadow} category foreground weighting is a crucial component for recovering a background model that is not corrupted by foreground objects. Figure~\ref{fig:fg-weighting_bungalows} showcases this effect. We found that foreground weighting allows for larger step sizes $t_{min}$ without producing ghost images.

\paragraph*{Step size} \quad The interplay between foreground weighting and different choices for the step size is displayed in Figure~\ref{fig:fg-weighting_fall}. Without any weighting the dynamic elements can be compensated using a large step size, but large foreground objects are incorporated too quickly into the background, which leads to reconstruction failure in some cases. A very small step size prohibits the formation of ghost images, but makes it impossible to adapt to the dynamic elements and lighting changes. Finally, combining a large step size and foreground weighting solves both problems.
\begin{figure}
\centering
  \subfigure[Frame 1291 of \textit{traffic} video and ground truth ]{
   \includegraphics[width=0.23\textwidth]{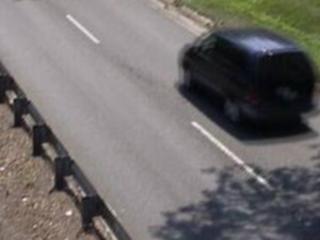}~
   \includegraphics[width=0.23\textwidth]{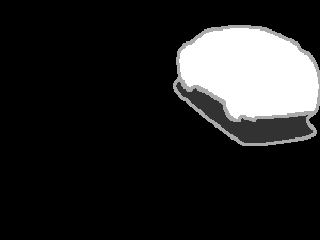}~
   }
   \subfigure[p=2.0  ]{
   \includegraphics[width=0.23\textwidth]{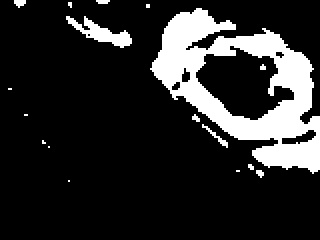}~
   \includegraphics[width=0.23\textwidth]{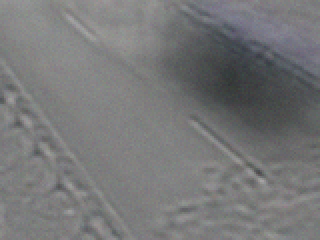}~
   }\\
   \subfigure[p=1.0 ]{
   \includegraphics[width=0.23\textwidth]{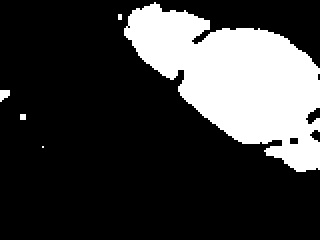}~
   \includegraphics[width=0.23\textwidth]{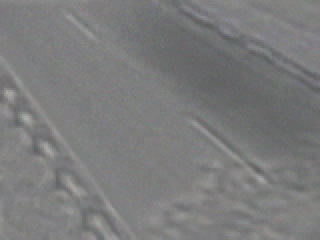}~

   }
   \subfigure[p=0.5 ]{
   \includegraphics[width=0.23\textwidth]{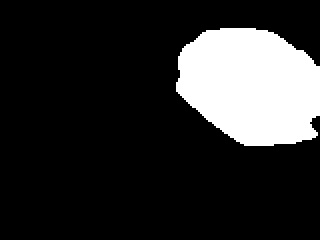}~
   \includegraphics[width=0.23\textwidth]{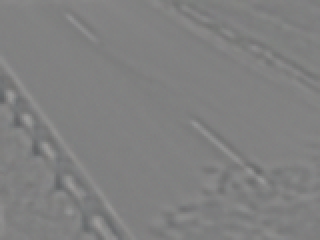}

   }
\caption{Effects of $\ell_p$-norm parameter $p$ on background model learning and fitting in jittery videos ($k=15$, $\mu=10^{-3}$, $\omega=1.0$)}
\label{fig:pchoice_traffic}       
\end{figure}
\begin{figure}
\centering
  \subfigure[Frame 1243 of \textit{traffic} video and ground truth ]{
   \includegraphics[width=0.23\textwidth]{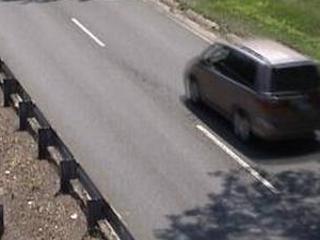}~
   \includegraphics[width=0.23\textwidth]{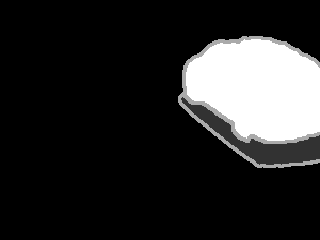}~

   }
   \subfigure[$\mu=10^{-1}$ ]{
   \includegraphics[width=0.23\textwidth]{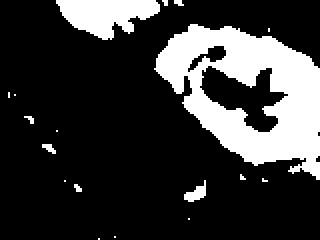}~
   \includegraphics[width=0.23\textwidth]{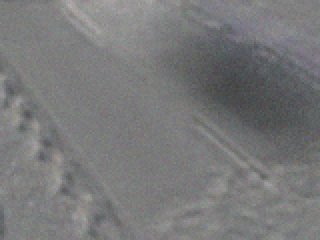}~

   }\\
   \subfigure[$\mu=10^{-3}$ ]{
   \includegraphics[width=0.23\textwidth]{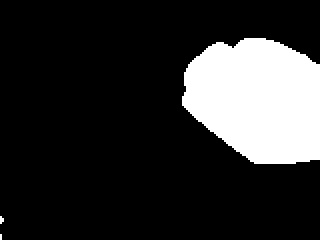}~
   \includegraphics[width=0.23\textwidth]{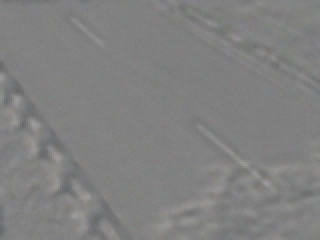}~

   }
   \subfigure[$\mu=10^{-5}$ ]{
   \includegraphics[width=0.23\textwidth]{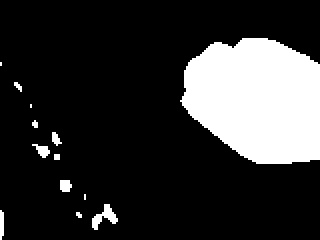}~
   \includegraphics[width=0.23\textwidth]{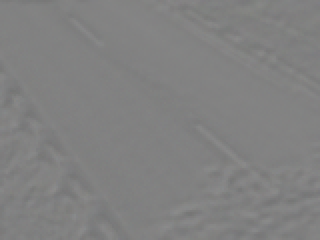}

   }
\caption{Effects of $\ell_p$-norm parameter $\mu$ on background model learning and fitting in jittery videos ($k=15$, $p=0.5$, $\omega=1.0$)}
\label{fig:mu_choice}       
\end{figure}
\paragraph*{Cost function parameters} \quad In contrast to the $\ell_1$-norm, the choice of $p$ and $\mu$ in the smoothed $\ell_p$-norm \eqref{eq:smoothlpnorm} offers control over the degree of robustness to outliers in the data. To analyze this effect, we extend the definition of the smoothed $\ell_p$-norm cost function to the case $p\geq 1$. Figure \ref{fig:pchoice_traffic} illustrates that lowering the parameter $p$ can reduce the rate of incorporating foreground objects into the background.

\begin{figure}
\begin{tikzpicture}
    \begin{axis}[
      xmode=log,
      ylabel=F-Measure,
      xlabel= $\mu$,
      xmin=0,xmax=0.2,
      ymin=0.65,ymax=0.85,
      legend style={at={(0.85,0.25)},anchor=north}
      ] 
        \addplot+[sharp plot,color=Diagram3,no markers,line width = 2,dashdotted] coordinates {
            (0.0056, 0.37) 
            (0.0112, 0.552) 
            (0.0225, 0.711) 
            (0.0338, 0.713) 
            (0.0450, 0.717)
            (0.0562, 0.716)
            (0.0675, 0.721)
            (0.09, 0.706)
            };
        \addplot+[sharp plot,color=Diagram5,no markers,line width = 2,dashed] coordinates {
            (0.0077, 0.49) 
            (0.0153, 0.675) 
            (0.0306, 0.735) 
            (0.0459, 0.741) 
            (0.0612, 0.751)
            (0.0760, 0.747)
            (0.0919, 0.748)
            (0.1225, 0.737)
            };
        \node[anchor=west] (source) at (axis cs:0.025,0.8)					{\textbullet\ Predicted optimal $\mu$ for $\delta=0.35$};
       \node (destination) at (axis cs:0.0919,0.748){};
       \draw[->](source)--(destination);

        \node[anchor=west] (source) at (axis cs:0.015,0.76)					{\textbullet\ Predicted optimal $\mu$ for $\delta=0.3$};
       \node (destination) at (axis cs:0.0675,0.721){};
       \draw[->](source)--(destination);       
       
        \legend{$\delta=0.3$, $\delta=0.35$}
    \end{axis}
    \label{fig:mu_opt}
    
\end{tikzpicture}

\begin{tikzpicture}
    \begin{axis}[
      xmode=log,
      ylabel=F-Measure,
      xlabel= $\mu$,
      xmin=0,xmax=0.2,
      ymin=0.7,ymax=0.9,
      legend style={at={(0.85,0.25)},anchor=north}
      ] 
        \addplot+[sharp plot,color=Diagram3,no markers,line width = 2,dashdotted] coordinates {
            (0.0056, 0.826) 
            (0.0112, 0.837) 
            (0.0225, 0.869) 
            (0.0338, 0.870756) 
            (0.0450, 0.873)
            (0.0562, 0.876)
            (0.0675, 0.878)
            (0.09, 0.873)
            };
        \addplot+[sharp plot,color=Diagram5,no markers,line width = 2,dashed] coordinates {
            (0.0077, 0.825238) 
            (0.0153, 0.841412) 
            (0.0306, 0.847554) 
            (0.0459, 0.853369) 
            (0.0612, 0.858658)
            (0.0760, 0.833419)
            (0.0919, 0.865304)
            (0.1225, 0.86388)
            };
        \node[anchor=west] (source) at (axis cs:0.015,0.77)					{\textbullet\ Predicted optimal $\mu$ for $\delta=0.35$};
       \node (destination) at (axis cs:0.0919,0.865304){};
       \draw[->](source)--(destination);

        \node[anchor=west] (source) at (axis cs:0.005,0.80)					{\textbullet\ Predicted optimal $\mu$ for $\delta=0.3$};
       \node (destination) at (axis cs:0.0675,0.878){};
       \draw[->](source)--(destination);       
       
        \legend{$\delta=0.3$, $\delta=0.35$}
    \end{axis}
    \label{fig:mu_opt2}
    
\end{tikzpicture}
\caption{Evaluation of $\mu$-heuristic \ref{eq:opt_mu2} for \textit{traffic} (top) and \textit{badminton} (bottom)}
\end{figure}
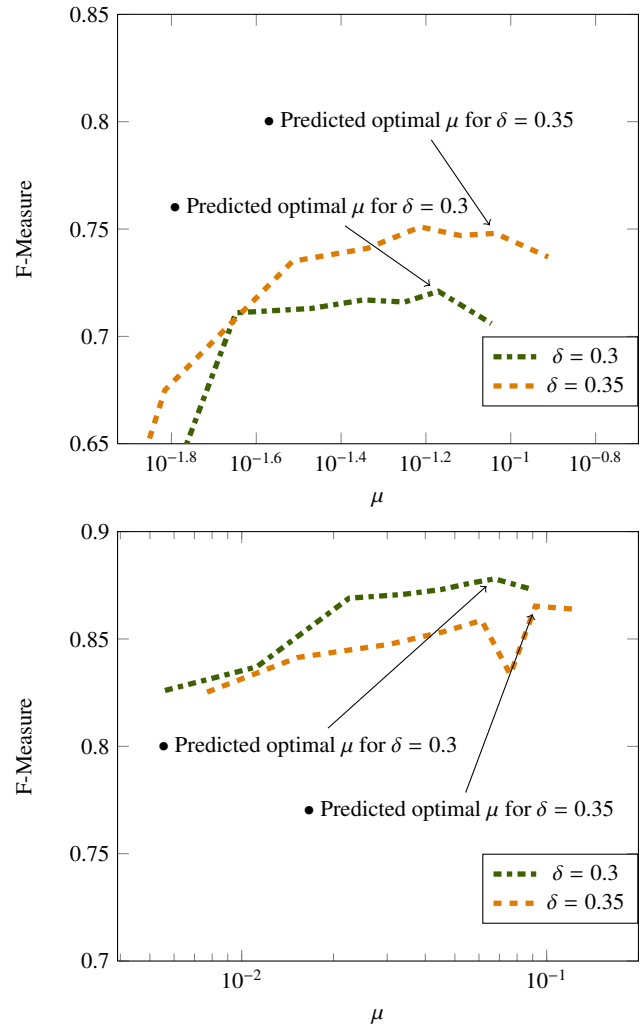

As discussed in Section~\ref{sec:rpcabgmodels}, Robust PCA algorithms aim for accurately recovering the assumed low-rank component $L$ of a data matrix $X$. However, when a decision is based on thresholding of $S=X-L$, a certain degree of reconstruction accuracy is sufficient and further decreasing the reconstruction error after it has fallen below the threshold does not add to the performance, but unnecessarily requires computational resources. In this sense, an estimate of $L$ should be considered sufficiently close to the true $L$ if it does not produce a false positive. To reflect this idea in the cost function, we require that starting at the threshold $\delta$ the partial derivative of the cost function with respect to the reconstruction error becomes smaller as we approach zero and becomes larger as we approach the threshold from above. This translates to the constraint
\begin{align}
	&& \frac{\partial ^2 h_{\mu}}{\partial r_i ^2}\bigg|_{r_i=\delta} = 0, \: \mbox{for fixed } p < 1 . \label{eq:opt_mu}
\end{align}
 A quick calculation reveals that
\begin{align}
	&& \mu=\delta ^2(1-p). \label{eq:opt_mu2}
\end{align}
meets this constraint. To back-up that this coupling between $\mu$ and $\delta$ indeed leads to near-optimal smoothing, we conduct an experiment in which we evaluate a number of smoothing parameters for two different thresholds on the \textit{traffic} and \textit{badminton} videos. The results in Figure~\ref{fig:mu_opt} confirm our theoretical analysis. Furthermore, the performance degrades markedly if a very small smoothing parameter is chosen.
\section{Conclusion}

We have presented a novel subspace tracking algorithm which combines concepts of previous methods and introduces a novel weighted robust cost function tailored to the task of background modelling and foreground segmentation from video data. The method is implemented on a GPU, achieves frame rates between 30 and 45 FPS on images in a resolution of $160¸\times 120$ and is thus real-time capable. One of the noteworthy features of the method is that it does not need a batch initialization phase, but learns the background model from corrupted streaming video. This has the advantage that no camera frames need to be stored at any time during operation.
 
pROST should be considered a basic building block for larger, more sophisticated systems for background subtraction. Future work will include the extension with a shadow detection mechanism and more sophisticated pre- and post-processing techniques. 

We have evaluated the algorithm on the \emph{changedetection.net} benchmark and show that it outperforms the conceptually similar GRASTA algorithm in many categories. Our method is particularly suitable for videos recorded by highly unstable cameras, ranking first in this category by a large margin, and it can thus be considered an advancement in this research area.


\bibliographystyle{spmpsci}      
\bibliography{bibliography}   

\end{document}